\documentclass[11pt,a4paper]{article}
\usepackage[hyperref]{acl2017}
\usepackage{times}
\usepackage{latexsym}
\usepackage{booktabs}
\usepackage{graphicx}
\usepackage[plain]{fancyref}

\usepackage{url}

\aclfinalcopy 

\title{SU-RUG at the CoNLL-SIGMORPHON 2017 shared task: \\Morphological Inflection with Attentional Sequence-to-Sequence Models}

\author{Robert \"Ostling \\
    Department of Linguistics \\
    Stockholm University \\
    Sweden \\
    {\tt robert@ling.su.se} \\\And
    Johannes Bjerva\thanks{$^*$This work was carried out while the second author was visiting the Department of Linguistics, Stockholm University.}\\
  Center for Language and Cognition Groningen \\
  University of Groningen \\
  The Netherlands \\
  {\tt j.bjerva@rug.nl} \\}

\date{}

\begin{document}
\maketitle
\begin{abstract}
    This paper describes the Stockholm University/University of Groningen
    (SU-RUG) system for the SIGMORPHON 2017 shared task on morphological
    inflection. Our system is based on an attentional sequence-to-sequence
    neural network model using Long Short-Term Memory (LSTM) cells, with
    joint training of morphological inflection and the inverse transformation,
    i.e.\ lemmatization and morphological analysis.
    Our system outperforms the baseline with a large margin, and our
    submission ranks as the $4_{th}$ best team for the track we participate in
    (task 1, high-resource).
\end{abstract}

\section{Introduction}

We focus on task 1 of the SIGMORPHON 2017 shared task
\citep{Cotterell2017sigmorphon}, morphological inflection. The task is to
learn the mapping from a lemma and morphological description to the
corresponding inflected form. For instance, the English verb lemma
\textit{torment} with the features \textsc{3.sg.prs} should be mapped to
\textit{torments}. As our model is poorly suited for low-resource conditions,
we only submitted results for the 51 languages
with high-resource training data available in the shared task (i.e., excluding Scottish Gaelic).

\section{Background}

The results of the SIGMORPHON 2016 shared task \citep{Cotterell2016sigmorphon}
indicated that the attentional sequence-to-sequence model of
\citet{Bahdanau2014nmt} is very suitable for this task
\citep{Kann2016sigmorphon}, so we use this framework as the basis of our
model.

A recent trend in neural machine translation is to use back-translated text
\citep{Sennrich2016backtranslation} as a way to benefit from additional
monolingual data in the target language. There is also work on translation
models with reconstruction loss, which encourages solutions that can be
translated back to their original \citep{Tu2016reconstruction}.  These
developments are technically similar to our semi-supervised training below.

\section{Method}
\label{sec:method}

\begin{figure*}[htbp]
    \centering
    \includegraphics[width=0.8\textwidth]{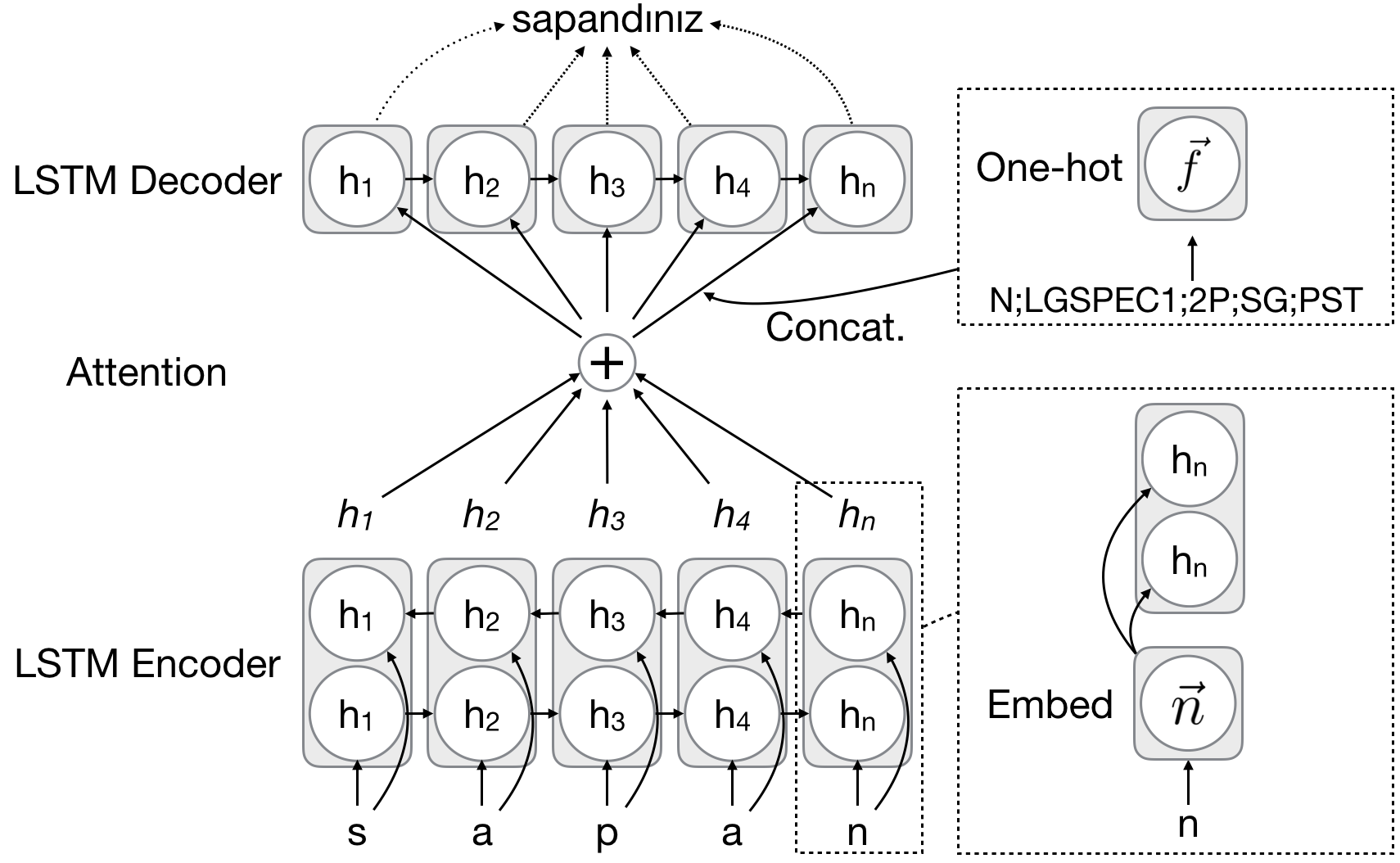}
    \caption{System architecture, consisting of an attentional sequence-to-sequence model with LSTMs.
    \label{fig:sigmorphon_system_arch}}
\end{figure*}

Our system is based on the attentional sequence-to-sequence model of
\citet{Bahdanau2014nmt} with Long Short-Term Memory (LSTM) cells
\citep{Hochreiter1997lstm} and variational dropout
\citet{Gal2016theoretically}. The main innovation is that our inflection model
is trained jointly with the reverse process, that is, lemmatization and
morphological analysis. This can be done in two ways:
\begin{enumerate}
    \item Fully supervised, where we simply train the forward (inflection) and
        backward (lemmatization and morphological analysis) model jointly with
        shared character embeddings.
    \item Semi-supervised, where supervised examples are mixed with examples
        where only the inflected target form is used. This form is passed
        first through the backward model, a greedy search to obtain a unique
        lemma, and finally through the forward model to reconstruct the
        inflected form.
\end{enumerate}
Our official submission only includes results from fully supervised training
(method 1), due to time constraints, but \Fref{sec:results} contains a
comparison between the two versions on the development set.
The system architecture is shown in Figure~\ref{fig:sigmorphon_system_arch}
for the forward (inflection) model. The backward (lemmatizer) model has
separate parameters, except the embeddings, but is structurally identical
except for two details: instead of passing the morphological feature
information to the decoder (via a single fully connected layer), we predict
the features from the final state of the encoder LSTM (via a separate fully
connected layer).

Our implementation is based on the Chainer library \citep{Tokui2015chainer}
and available at \url{github.com/bjerva/sigmorphon2017}.

\section{Model configuration}

For the official submission, we use 128 LSTM cells for the (unidirectional)
encoder, decoder, attention mechanism, character embeddings,
 as well as for the fully connected
layers for morphological features encoding/prediciton.  We use a dropout
factor of 0.5 throughout the network, including the recurrent parts.  For
optimization, we use Adam \citep{Kingma2014adam} with default parameters.
Each model is trained for 48 hours on a single CPU, using a batch-size of 64,
and the model parameters during this time that give the lowest development set mean Levenshtein
distance are saved.  For the official submission, we used an ensemble of two
such models, using a beam search of width 10 to select the final inflection
candidate.

\section{Results and Analysis}
\label{sec:results}

The system has high performance
in general, with a macro-average accuracy of 93.6\%, and edit distance of
0.14. This is substantially higher than the baseline (77.8\% accuracy and 0.5 edit distance),
and ranks as the $9_{th}$ best run, and $4_{th}$ best team in this SIGMORPHON 2017 shared task setting.
Furthermore, the difference in scores between our run and the best run overall is low ($1.75\%$ accuracy and $0.04$ edit distance).
Table~\ref{tab:results} contains a detailed version of the official results our system on the shared
task, in the \textit{high} setting of Task 1.


\renewcommand{\arraystretch}{0.882} 
\begin{table}[htbp]
    \caption{Our system's official results on the SIGMORPHON-2017 shared task-1 test set in the \textit{high} setting.
    \vspace{0.1cm} 
    \label{tab:results}}
    \centering
    \begin{tabular}{lrr}
        \toprule
        \textbf{Language} & \textbf{Accuracy} & \textbf{Edit dist.} \\
        \midrule
        Albanian & 97.9 & 0.07 \\
        Arabic & 89.8 & 0.39 \\
        Armenian & 95.6 & 0.08 \\
        Basque & 100.0 & 0.00 \\
        Bengali & 99.0 & 0.05 \\
        Bulgarian & 96.7 & 0.07 \\
        Catalan & 97.8 & 0.06 \\
        Czech & 92.0 & 0.15 \\
        Danish & 93.8 & 0.09 \\
        Dutch & 95.9 & 0.07 \\
        English & 96.6 & 0.07 \\
        Estonian & 96.8 & 0.08 \\
        Faroese & 84.6 & 0.31 \\
        Finnish & 91.0 & 0.17 \\
        French & 87.5 & 0.24 \\
        Georgian & 97.6 & 0.05 \\
        German & 89.5 & 0.21 \\
        Haida & 95.0 & 0.10 \\
        Hebrew & 99.0 & 0.01 \\
        Hindi & 99.8 & 0.00 \\
        Hungarian & 84.8 & 0.35 \\
        Icelandic & 86.3 & 0.25 \\
        Irish & 87.6 & 0.35 \\
        Italian & 96.8 & 0.09 \\
        Khaling & 98.3 & 0.03 \\
        Kurmanji & 93.8 & 0.10 \\
        Latin & 75.3 & 0.39 \\
        Latvian & 95.4 & 0.08 \\
        Lithuanian & 91.0 & 0.15 \\
        Lower Sorbian & 96.9 & 0.06 \\
        Macedonian & 96.6 & 0.06 \\
        Navajo & 88.9 & 0.28 \\
        Northern Sami & 94.5 & 0.12 \\
        Norwegian (Bokm{\aa}l) & 92.4 & 0.13 \\
        Norwegian (Nynorsk) & 89.4 & 0.18 \\
        Persian & 99.3 & 0.01 \\
        Polish & 90.6 & 0.22 \\
        Portuguese & 98.8 & 0.02 \\
        Quechua & 100.0 & 0.00 \\
        Romanian & 86.4 & 0.42 \\
        Russian & 89.3 & 0.31 \\
        Serbo-Croatian & 90.1 & 0.24 \\
        Slovak & 93.1 & 0.13 \\
        Slovene & 96.6 & 0.07 \\
        Sorani & 88.6 & 0.14 \\
        Spanish & 93.5 & 0.15 \\
        Swedish & 91.8 & 0.13 \\
        Turkish & 96.6 & 0.11 \\
        Ukrainian & 94.2 & 0.11 \\
        Urdu & 99.7 & 0.01 \\
        Welsh & 99.0 & 0.03 \\
        \midrule
        \textbf{Average} & \textbf{93.6}  & \textbf{0.14} \\
        \bottomrule
    \end{tabular}
\end{table}

\renewcommand{\arraystretch}{0.8797} 
\begin{table}[htbp]
    \caption{Our system's result on the SIGMORPHON-2017 shared task-1
        development set, comparing fully supervised training (\textbf{Full})
        to our semi-supervised method (\textbf{Semi}).}
    \vspace{0.1cm} 
    \label{tab:results-semi}
    \centering
    \begin{tabular}{lrr}
        \toprule
         & \multicolumn{2}{c}{\bf Accuracy} \\
        \textbf{Language} & \textbf{Full} & \textbf{Semi} \\
        \midrule
        Albanian & 97.6 & 97.0 \\
        Arabic & 93.0 & 93.1 \\
        Armenian & 96.9 & 97.1 \\
        Basque & 99.0 & 99.0 \\
        Bengali & 99.0 & 99.0 \\
        Bulgarian & 95.8 & 96.0 \\
        Catalan & 98.0 & 98.3 \\
        Czech & 92.5 & 93.1 \\
        Danish & 95.8 & 95.9 \\
        Dutch & 96.8 & 97.1 \\
        English & 96.6 & 96.3 \\
        Estonian & 97.4 & 97.6 \\
        Faroese & 86.7 & 87.1 \\
        Finnish & 91.2 & 91.4 \\
        French & 89.8 & 89.3 \\
        Georgian & 97.9 & 97.9 \\
        German & 87.8 & 89.6 \\
        Hebrew & 98.8 & 98.7 \\
        Hindi & 99.9 & 99.8 \\
        Hungarian & 86.8 & 87.1 \\
        Icelandic & 88.1 & 88.6 \\
        Irish & 89.0 & 89.5 \\
        Italian & 97.0 & 97.2 \\
        Kurmanji & 92.4 & 92.7 \\
        Latin & 75.6 & 75.9 \\
        Latvian & 95.2 & 96.4 \\
        Lithuanian & 90.3 & 89.6 \\
        Lower Sorbian & 97.7 & 96.3 \\
        Macedonian & 95.3 & 95.0 \\
        Navajo & 88.2 & 85.2 \\
        Northern Sami & 94.4 & 93.5 \\
        Norwegian (Bokm{\aa}l) & 91.8 & 92.7 \\
        Norwegian (Nynorsk) & 92.3 & 92.4 \\
        Persian & 99.5 & 99.6 \\
        Polish & 91.0 & 92.0 \\
        Portuguese & 98.6 & 98.0 \\
        Quechua & 100.0 & 100.0 \\
        Romanian & 87.4 & 88.2 \\
        Russian & 89.8 & 88.1 \\
        Serbo-Croatian & 89.5 & 89.7 \\
        Slovak & 95.2 & 94.8 \\
        Slovene & 96.7 & 97.0 \\
        Sorani & 90.9 & 90.3 \\
        Spanish & 94.3 & 95.7 \\
        Swedish & 90.9 & 90.1 \\
        Turkish & 97.5 & 97.2 \\
        Ukrainian & 94.0 & 92.7 \\
        Urdu & 99.5 & 99.2 \\
        Welsh & 100.0 & 100.0 \\
        \midrule
        \textbf{Average} & \textbf{93.9}  & \textbf{93.8} \\
        \bottomrule
    \end{tabular}
\end{table}

Notably, the system has an accuracy of 100\% on both Basque and
Quechua, which indicates that it is capable of fully learning the rules of
very regular morphological systems. The relatively high accuracy on Semitic
languages (Arabic: 89.8\%, Hebrew: 99.0\%) again confirms the ability of
encoder-decoder models to also handle non-concatenative morphology.

Latin has the lowest accuracy by far, and the reason seems to be that the
provided shared task data lacks vowel length distinctions in the lemma but
uses them in the inflected forms. This missing lexical information is
difficult to predict accurately. Evaluating with vowel length distinctions
gives an accuracy of 75.6\% (Latin development set), compared to 91.5\%
without. The latter accuracy score is in line with other Romance languages (French
90.8\%, Spanish 94.3\%, Italian 97.0\%).


We also investigated whether the semi-supervised approach described in
\Fref{sec:method} has any effect on accuracy. The results on the development set,
 presented in
\Fref{tab:results-semi}, indicate that there is no systematic effect (the
macro-averaged accuracy drops marginally from 93.9\% to 93.8\%).

\section{Conclusions}

We implemented a system using an attentional sequence-to-sequence model with Long Short-Term Memory (LSTM) cells.
As our model is poorly suited for low-resource conditions, we only participated in the high-resource setting.
Our inflection model is trained jointly with the reverse process, that is, lemmatization and
morphological analysis.
The system significantly outperforms the baseline system, and performs well compared to other submitted systems,
showing that this approach is very suitable for morphological inflection, given sufficient amounts of data.

\section*{Acknowledgments}
The authors would like to thank the reviewers, and Johan Sjons for their comments on previous versions of this manuscript.
This work was partially funded by the NWO--VICI grant \mbox{``Lost in Translation -- Found in Meaning''} (288-89-003).
This work was performed on the Abel Cluster, owned by the University of Oslo and the Norwegian metacenter
for High Performance Computing (NOTUR), and operated by the Department for Research Computing at USIT,
the University of Oslo IT-department.

\bibliographystyle{acl_natbib}
\bibliography{sigmorphon2017}

\begin{thebibliography}{}
\expandafter\ifx\csname natexlab\endcsname\relax\def\natexlab#1{#1}\fi

\bibitem[{Bahdanau et~al.(2014)Bahdanau, Cho, and Bengio}]{Bahdanau2014nmt}
Dzmitry Bahdanau, Kyunghyun Cho, and Yoshua Bengio. 2014.
\newblock Neural machine translation by jointly learning to align and
  translate.
\newblock {\em CoRR\/} abs/1409.0473.

\bibitem[{Cotterell et~al.(2017)Cotterell, Kirov, Sylak-Glassman, Walther,
  Vylomova, Xia, Faruqui, K{\"u}bler, Yarowsky, Eisner, and
  Hulden}]{Cotterell2017sigmorphon}
Ryan Cotterell, Christo Kirov, John Sylak-Glassman, G{\'e}raldine Walther,
  Ekaterina Vylomova, Patrick Xia, Manaal Faruqui, Sandra K{\"u}bler, David
  Yarowsky, Jason Eisner, and Mans Hulden. 2017.
\newblock The {CoNLL-SIGMORPHON} 2017 shared task: Universal morphological
  reinflection in 52 languages.
\newblock In {\em Proceedings of the CoNLL-SIGMORPHON 2017 Shared Task:
  Universal Morphological Reinflection\/}. Association for Computational
  Linguistics, Vancouver, Canada.

\bibitem[{Cotterell et~al.(2016)Cotterell, Kirov, Sylak-Glassman, Yarowsky,
  Eisner, and Hulden}]{Cotterell2016sigmorphon}
Ryan Cotterell, Christo Kirov, John Sylak-Glassman, David Yarowsky, Jason
  Eisner, and Mans Hulden. 2016.
\newblock The sigmorphon 2016 shared task: Morphological reinflection.
\newblock In {\em Proceedings of the 14th SIGMORPHON Workshop on Computational
  Research in Phonetics, Phonology, and Morphology\/}. Association for
  Computational Linguistics, Berlin, Germany, pages 10--22.

\bibitem[{Gal and Ghahramani(2016)}]{Gal2016theoretically}
Yarin Gal and Zoubin Ghahramani. 2016.
\newblock A theoretically grounded application of dropout in recurrent neural
  networks.
\newblock In {\em Advances in Neural Information Processing Systems 29
  (NIPS)\/}.

\bibitem[{Hochreiter and Schmidhuber(1997)}]{Hochreiter1997lstm}
Sepp Hochreiter and J{\"u}rgen Schmidhuber. 1997.
\newblock Long short-term memory.
\newblock {\em Neural Computation\/} 9(8):1735--1780.

\bibitem[{Kann and Sch\"{u}tze(2016)}]{Kann2016sigmorphon}
Katharina Kann and Hinrich Sch\"{u}tze. 2016.
\newblock Med: The lmu system for the sigmorphon 2016 shared task on
  morphological reinflection.
\newblock In {\em Proceedings of the 14th SIGMORPHON Workshop on Computational
  Research in Phonetics, Phonology, and Morphology\/}. Association for
  Computational Linguistics, Berlin, Germany, pages 62--70.

\bibitem[{Kingma and Ba(2015)}]{Kingma2014adam}
Diederik~P. Kingma and Jimmy Ba. 2015.
\newblock Adam: {A} method for stochastic optimization.
\newblock In {\em The International Conference on Learning Representations\/}.

\bibitem[{Sennrich et~al.(2016)Sennrich, Haddow, and
  Birch}]{Sennrich2016backtranslation}
Rico Sennrich, Barry Haddow, and Alexandra Birch. 2016.
\newblock Improving neural machine translation models with monolingual data.
\newblock In {\em Proceedings of the 54th Annual Meeting of the Association for
  Computational Linguistics (Volume 1: Long Papers)\/}. Association for
  Computational Linguistics, Berlin, Germany, pages 86--96.

\bibitem[{Tokui et~al.(2015)Tokui, Oono, Hido, and Clayton}]{Tokui2015chainer}
Seiya Tokui, Kenta Oono, Shohei Hido, and Justin Clayton. 2015.
\newblock Chainer: a next-generation open source framework for deep learning.
\newblock In {\em Proceedings of Workshop on Machine Learning Systems
  (LearningSys) in The Twenty-ninth Annual Conference on Neural Information
  Processing Systems (NIPS)\/}.

\bibitem[{Tu et~al.(2016)Tu, Liu, Shang, Liu, and Li}]{Tu2016reconstruction}
Zhaopeng Tu, Yang Liu, Lifeng Shang, Xiaohua Liu, and Hang Li. 2016.
\newblock Neural machine translation with reconstruction.
\newblock {\em CoRR\/} abs/1611.01874.

\end{thebibliography}

\end{document}